\documentclass[conference]{IEEEtran}
\IEEEoverridecommandlockouts

\usepackage{cite}
\usepackage{amsmath,amssymb,amsfonts}
\usepackage{graphicx}
\usepackage{textcomp}
\usepackage[table]{xcolor}
\definecolor{bestrow}{gray}{0.90}
\definecolor{headrow}{gray}{0.85}
\definecolor{bestblue}{RGB}{0,64,190}
\newcommand{\best}[1]{\textcolor{bestblue}{\underline{\textbf{#1}}}}
\usepackage{booktabs}
\usepackage{multirow}
\usepackage{subcaption}
\usepackage{stfloats}

\begin{document}

\title{GRIPNet: Gaussian Radial Intensity Prior Guided Architecture for Pulmonary Nodule Detection in CT}

\author{Haojie Yang, Ran Su$^*$\\
College of Intelligence and Computing, Tianjin University, Tianjin, China\\
\texttt{haojae@tju.edu.cn}, \texttt{ran.su@tju.edu.cn}\\
$^*$Corresponding author}

\maketitle

\begin{abstract}
Lung cancer causes more deaths than any other malignancy, and low-dose CT screening is the main pathway to early diagnosis. That pathway hinges on the smallest lesions, yet nodules below six millimeters remain hard to detect, because most methods treat a nodule as a generic object and ignore the imaging physics behind its appearance. We show that this appearance is highly regular. Intensity peaks at the geometric center of a nodule and decays radially in a Gaussian pattern, and a fit to 18,218 annotated lesions from three public benchmarks yields a mean radial coefficient of determination above 0.86 in every dataset and size stratum. A square convolution samples both axes uniformly and is mismatched to this radial signal, most severely for small nodules. Guided by this evidence, we propose GRIPNet (Gaussian Radial Intensity Prior Network), a detector in which every module maps to a measurable property of the intensity distribution. Pinwheel convolutions decompose radial gradients, a dual-frequency module separates boundary detail from structural context, dilated masked attention matches the decay extent, and an adaptive loss reweights samples by conspicuity. GRIPNet raises mAP@0.5 to 95.3, 91.6 and 97.9 percent on KanserSet, LUNA16 and Lung-PET-CT-Dx while sharpening high-IoU localization at real-time speed.
\end{abstract}


\section{Introduction}
Lung cancer remains the leading cause of cancer death worldwide~\cite{sung2021global}. Finding it early multiplies the five-year survival rate, and low-dose CT screening has become the standard route to that early diagnosis~\cite{afridi2025minimally}. Deep detectors have improved steadily over the past decade~\cite{terven2023comprehensive,litjens2017survey}. Even so, micro-nodules below 6\,mm are still detected at only 65 to 75 percent sensitivity in routine reading, and ground-glass opacities with faint margins continue to confuse conventional feature extractors. The problem is structural rather than incidental. Most detectors treat a pulmonary nodule as a generic object and disregard the physical process that shapes its CT appearance~\cite{halder2020lung}, so they spend capacity relearning what a simple imaging prior would already provide.

\begin{figure}[t]
  \centering
  \includegraphics[width=0.92\linewidth]{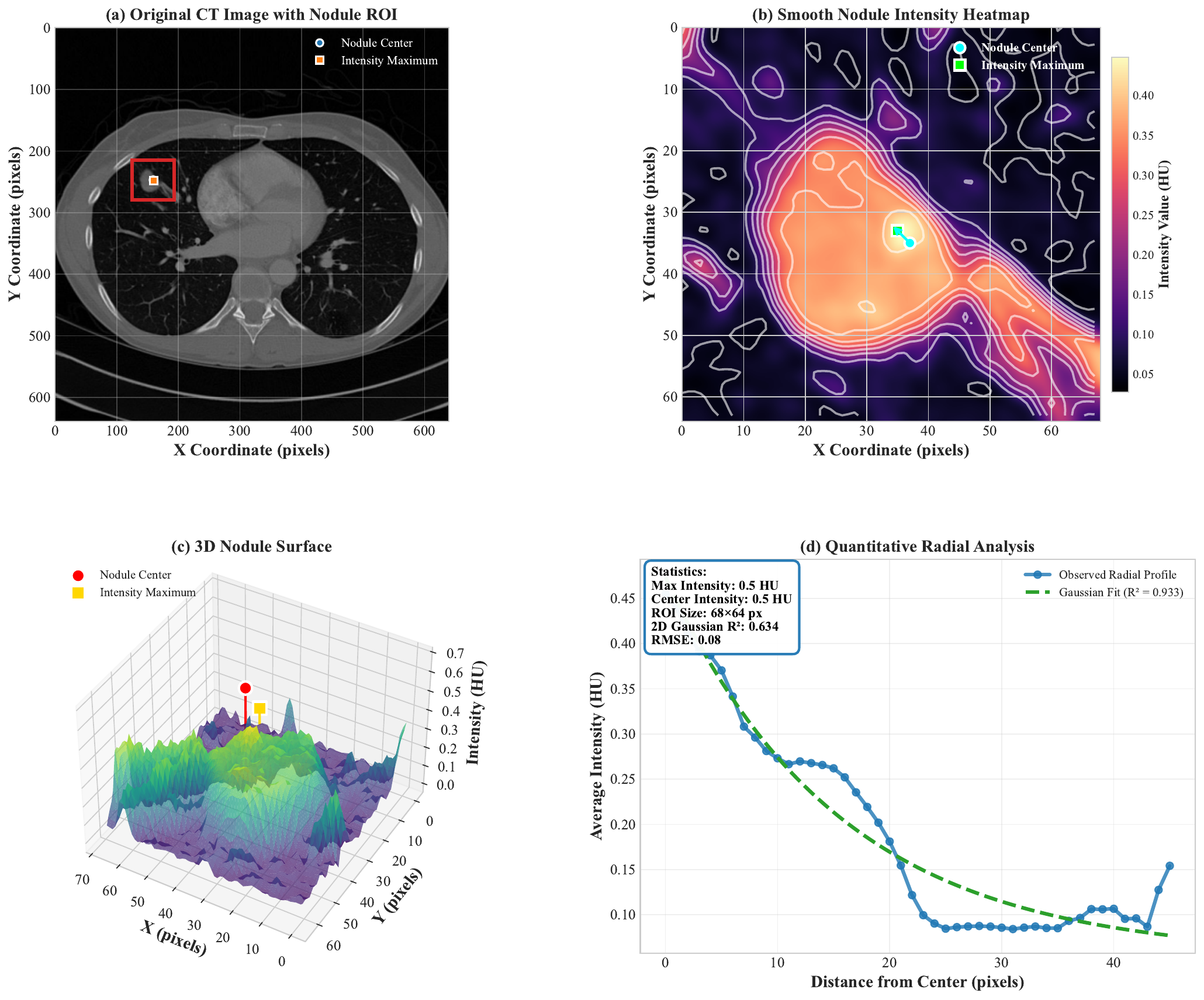}
  \caption{Gaussian-like radial intensity profile of a pulmonary nodule. Peak HU values concentrate at the geometric center and attenuate radially.}
  \label{fig:gaussian}
\end{figure}

A pulmonary nodule is far from a structureless region. Its CT appearance follows the partial-volume effect acting on a roughly spherical mass embedded in aerated parenchyma. Peak Hounsfield values concentrate at the geometric center and fade outward, as Fig.~\ref{fig:gaussian} shows. When a Gaussian is fitted to the radial profile, the mean coefficient of determination exceeds 0.86 across three public benchmarks. A conventional square kernel samples both axes evenly. It is therefore geometrically mismatched to these radial gradients and spends capacity on directions that carry little signal, and the mismatch hurts most on small nodules.

We propose GRIPNet, a Gaussian Radial Intensity Prior guided network in which every design choice traces to a quantifiable property of the nodule intensity distribution. Asymmetric pinwheel convolutions~\cite{yang2025pinwheel} decompose radial gradients along orthogonal axes. A dual-frequency module separates boundary detail from structural context. Dilated masked attention calibrates its receptive field to the Gaussian decay extent. An adaptive loss evolves its difficulty threshold as training progresses. These components form a coherent pipeline on a YOLOv11 backbone~\cite{yolov11_github} that reaches state-of-the-art accuracy at real-time speed.

The paper offers three contributions. We show that pulmonary nodule intensity profiles obey a Gaussian radial model, with a mean radial $R^2$ above 0.86 on three large-scale CT datasets. On that evidence we build GRIPNet, in which every module derives from a measurable physical property rather than from a generic architecture search. The resulting detector improves consistently over strong baselines, including recent lung-nodule-specific methods, and reaches a mAP@0.75 of 89.2\% on KanserSet at real-time throughput.

\section{Related Work}
Nodule detection moved from handcrafted descriptors to deep networks that learn features from data~\cite{halder2020lung}. Two-stage detectors reach high accuracy at heavy computational cost~\cite{ren2016faster}, while single-stage YOLO detectors trade a little accuracy for real-time speed and now dominate screening pipelines~\cite{terven2023comprehensive}. Recent lung-specific variants push small-object sensitivity through wider receptive fields in MSDet~\cite{cai2024msdet}, spatial-SE attention with an aspect-ratio penalty in the improved YOLOv11 of Song et al.~\cite{song2025improved}, and attention with atrous pooling in YOLOv5-CASP~\cite{zhang2023lung}. All refine where the network looks, yet none model why a nodule looks the way it does, so its radiometric structure never enters the architecture as a prior.

Multi-scale fusion and attention form a second thread. Feature pyramids merge information across resolutions~\cite{lin2017feature}, and spatial~\cite{woo2018cbam} or multi-scale attention~\cite{ouyang2023efficient} refines it. Vision Transformers raise medical accuracy, but their cost grows quadratically with token count and they reason in the spatial domain alone~\cite{dosovitskiy2020image,liu2021swin}. Our dual-frequency decomposition instead splits smooth interior intensity from fine boundary detail, adapting the blind-spot idea from self-supervised denoising~\cite{li2025rethinking}. Loss design matters just as much. Focal Loss rebalances hard and easy samples~\cite{lin2017focal}, and WiseIoU refines box regression for objects with a well-defined center~\cite{tong2023wise}. A prior drawn from the physics of the problem shrinks the hypothesis space and curbs overfitting, as physics-informed networks show~\cite{raissi2019physics}, which matters most where annotated nodules are scarce. GRIPNet applies this with a Gaussian radial prior measured from data rather than assumed, so every module rests on an empirical fact.

\section{Gaussian Radial Intensity Prior}
\label{sec:prior}
We first establish the empirical foundation. For each annotated nodule we extract a square region centered on the bounding-box centroid with side length 1.5 times the annotated diameter. The radial profile $I(r)$ averages HU values over concentric annuli,
\begin{equation}
\begin{aligned}
I(r) &= \frac{1}{|\mathcal{A}_r|}\sum_{(x,y)\in \mathcal{A}_r} I(x,y), \\
\mathcal{A}_r &= \{(x,y): r \leq \|(x,y)-\mathbf{c}\|_2 < r{+}1\},
\end{aligned}
\label{eq:radial}
\end{equation}
where $\mathbf{c}$ is the centroid. A Gaussian model is fitted by nonlinear least squares,
\begin{equation}
\hat{I}(r) = A\exp\!\left(-\frac{r^2}{2\sigma^2}\right) + B,
\label{eq:gauss}
\end{equation}
yielding amplitude $A$, spread $\sigma$ and coefficient of determination $R^2$.

We apply this protocol to KanserSet~\cite{model_api_dataset}, LUNA16~\cite{setio2017luna16} and Lung-PET-CT-Dx~\cite{li2020lungpetctdx}. As Table~\ref{tab:gauss} reports, the mean radial $R^2$ exceeds 0.86 in every stratum and more than 89\% of nodules pass 0.70, a threshold that indicates a good fit~\cite{motulsky2004fitting}. The spread $\bar{\sigma}$ scales with diameter from about 3 pixels for sub-6\,mm nodules to 12 pixels for those beyond 15\,mm. The regularity follows from CT physics: a dense core exceeds surrounding parenchyma by 150 to 300 HU, and partial-volume averaging produces intermediate values that fall with distance from the center~\cite{yankelevitz1999small}.

\begin{table*}[t]
\centering
\caption{Gaussian radial intensity fitting statistics across three benchmarks. $R^2_{\mathrm{radial}}$ is the coefficient of determination of the radial profile fit, $\bar\sigma$ is the mean spread in pixels, and the last column is the fraction of nodules with $R^2>0.70$.}
\label{tab:gauss}
\setlength{\tabcolsep}{7pt}
\small
\begin{tabular}{@{}llccccc@{}}
\toprule
\rowcolor{headrow}
\textbf{Dataset} & \textbf{Diameter / Subtype} & $N$ & $R^2_{\mathrm{radial}}$ (mean$\pm$std) & $\bar\sigma$ (px) & \textbf{Peak HU} & \textbf{\% $R^2{>}0.70$} \\
\midrule
\multirow{3}{*}{KanserSet}
& $<$6\,mm      & 412  & 0.891$\pm$0.062 & 3.2  & 185$\pm$41 & 94.2\% \\
& 6--15\,mm     & 2518 & 0.933$\pm$0.041 & 6.8  & 231$\pm$53 & 97.8\% \\
& $>$15\,mm     & 807  & 0.908$\pm$0.057 & 12.4 & 267$\pm$62 & 95.1\% \\
\midrule
\multirow{3}{*}{LUNA16}
& $<$6\,mm      & 298  & 0.862$\pm$0.078 & 2.8  & 172$\pm$38 & 89.6\% \\
& 6--15\,mm     & 641  & 0.921$\pm$0.048 & 5.9  & 218$\pm$49 & 96.4\% \\
& $>$15\,mm     & 247  & 0.897$\pm$0.061 & 11.1 & 254$\pm$58 & 93.5\% \\
\midrule
\multirow{4}{*}{Lung-PET-CT-Dx}
& Adenocarcinoma & 3864 & 0.873$\pm$0.071 & 8.7 & 198$\pm$46 & 91.3\% \\
& Small cell     & 3022 & 0.912$\pm$0.052 & 7.4 & 241$\pm$52 & 95.8\% \\
& Large cell     & 3015 & 0.924$\pm$0.046 & 9.2 & 258$\pm$57 & 96.7\% \\
& Squamous cell  & 3394 & 0.918$\pm$0.049 & 8.1 & 249$\pm$54 & 96.1\% \\
\bottomrule
\end{tabular}
\end{table*}

The prior carries three design implications that we build into GRIPNet. Radial decay generates gradients along every direction from the center, and a symmetric kernel cannot preferentially amplify the radial component, so an asymmetric operator captures these gradients with fewer parameters. The smooth Gaussian envelope occupies the low-frequency band while boundary irregularities carry high-frequency content, so an explicit frequency split lets two pathways specialize. The $3\sigma$ envelope spans roughly 9 to 36 pixels, so an attention window matched to that range integrates the full decay pattern without diluting the signal.

\section{Prior-Guided Architecture}
\label{sec:method}
\begin{figure*}[t]
  \centering
  \includegraphics[width=0.96\linewidth]{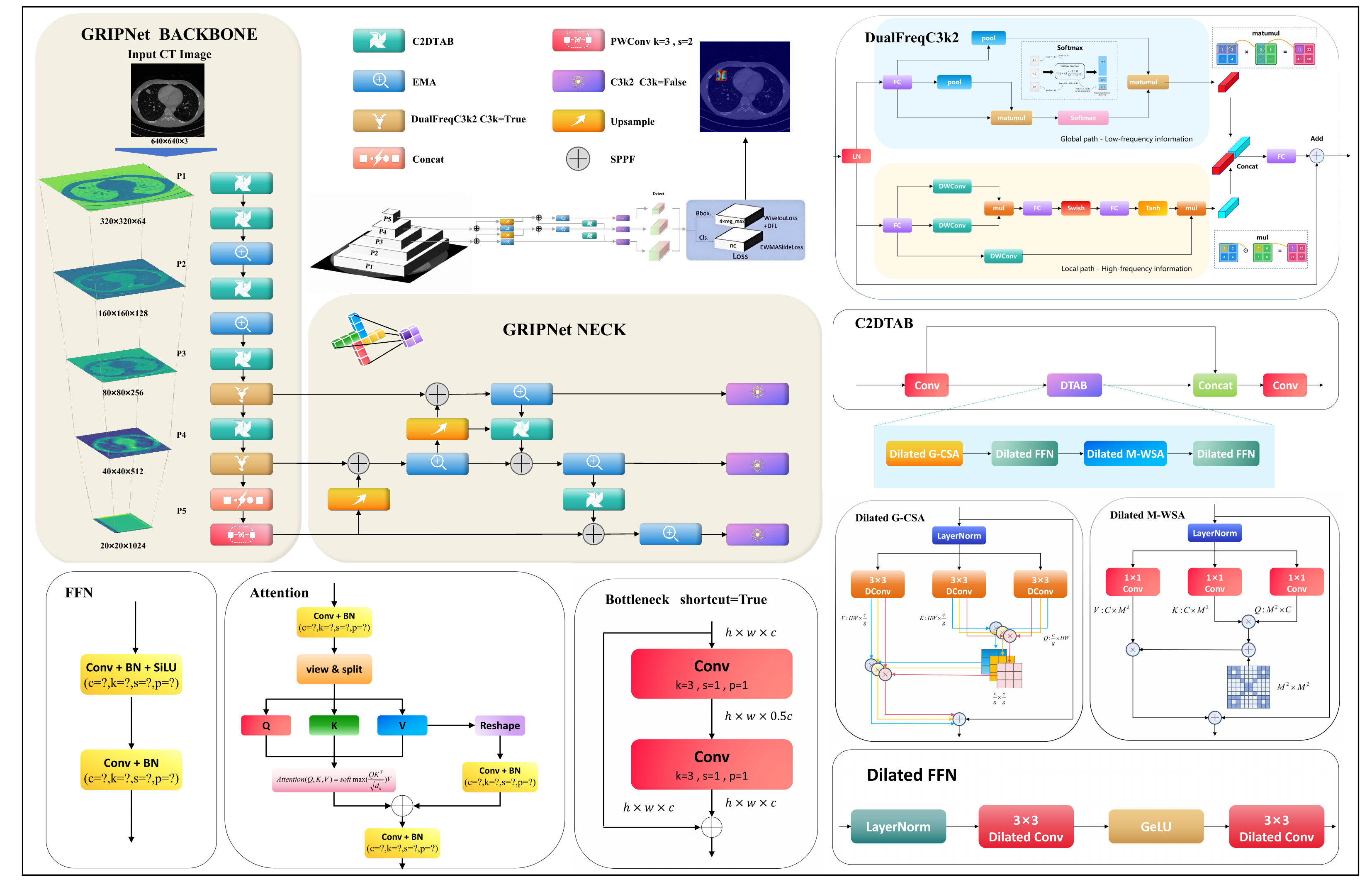}
  \caption{GRIPNet architecture. PWConv replaces symmetric downsampling to capture radial gradients. DualFreqC3k2 at P4 and P5 splits features into high-frequency morphology and low-frequency context. C2DTAB fuses scales with dilated masked attention matched to the Gaussian decay range. EMA-enhanced heads at P3 to P5 give scale-adaptive prediction.}
  \label{fig:arch}
\end{figure*}

GRIPNet translates the three implications into four modules on a YOLOv11 backbone that processes CT slices through progressive downsampling with fusion at levels P3 to P5, as shown in Fig.~\ref{fig:arch}.

\subsection{Pinwheel Convolution for Radial Gradients}
\label{sec:pwconv}
\begin{figure}[t]
  \centering
  \includegraphics[width=0.98\linewidth]{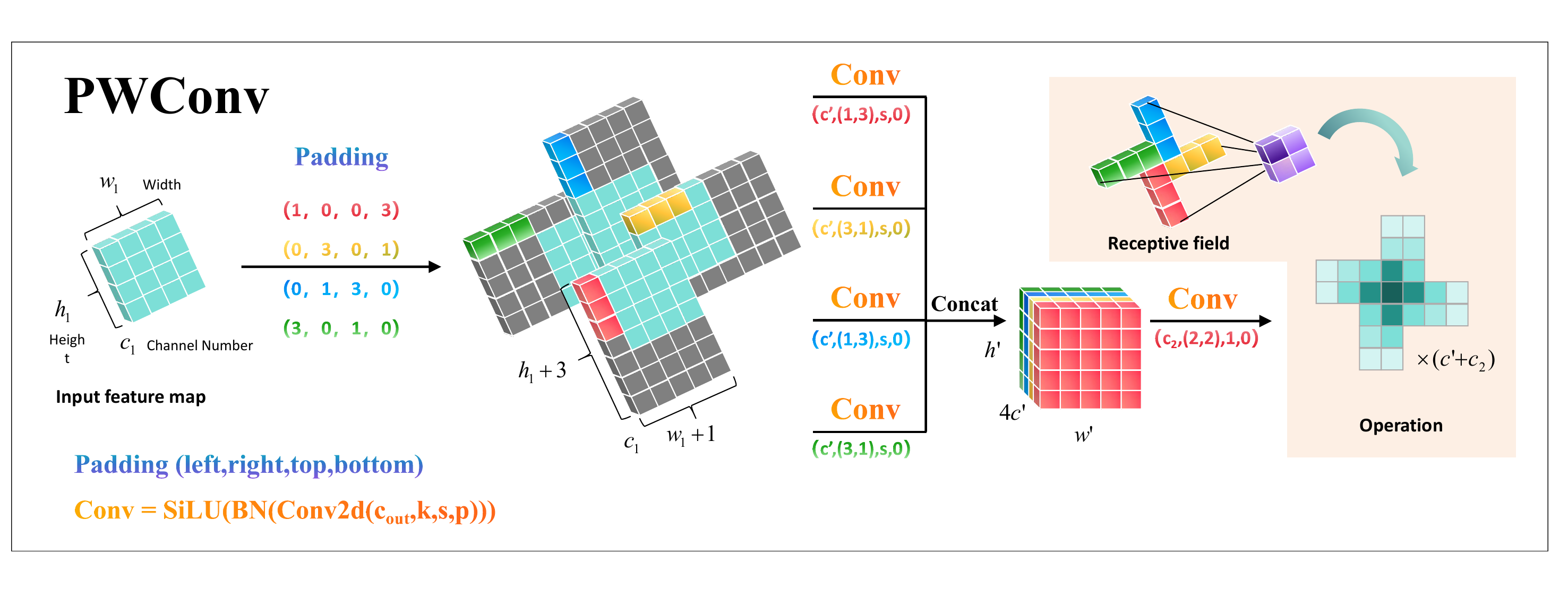}
  \caption{PWConv. Four asymmetric branches with directional padding capture orthogonal gradient components through $1\times3$ and $3\times1$ kernels, followed by $2\times2$ refinement.}
  \label{fig:pwconv}
\end{figure}

The prior implies that the most informative signal at a nodule boundary lies along the radial direction. A standard $3\times3$ convolution cannot amplify radial gradients without learning axis-specific filters that overfit on small medical datasets. Pinwheel Convolution~\cite{yang2025pinwheel} instead uses four asymmetric branches with directional padding. Given input $X$, the branches apply padding $\mathrm{Pad}_1{=}[3,0,1,0]$, $\mathrm{Pad}_2{=}[0,3,1,0]$, $\mathrm{Pad}_3{=}[1,0,3,0]$ and $\mathrm{Pad}_4{=}[0,1,0,3]$ followed by asymmetric kernels,
\begin{equation}
X'_i = \mathrm{SiLU}\!\left(\mathrm{BN}\!\left(W_i \otimes \mathrm{Pad}_i(X)\right)\right), \quad i\in\{1,2,3,4\},
\label{eq:pw1}
\end{equation}
where $W_1,W_3$ are $1\times3$ horizontal kernels and $W_2,W_4$ are $3\times1$ vertical kernels. The four outputs are concatenated and refined by a $2\times2$ convolution,
\begin{equation}
Y = \mathrm{SiLU}\!\left(\mathrm{BN}\!\left(W_{2\times2}\otimes \mathrm{Concat}[X'_1,X'_2,X'_3,X'_4]\right)\right).
\label{eq:pw2}
\end{equation}
A radial gradient projects onto horizontal and vertical components, and the four branches in Fig.~\ref{fig:pwconv} reconstruct the full radial field while using 22\% fewer parameters than a $3\times3$ convolution and expanding the receptive field by 177\%. We place PWConv at the stride-2 downsampling stages.

\subsection{Dual-Frequency Feature Decomposition}
\label{sec:dualfreq}
The Gaussian envelope lives in the low-frequency band, whereas spiculations, lobulations and calcifications carry diagnostic high-frequency detail. Routing both through a single pathway forces the network to balance competing objectives. DualFreqC3k2 resolves this with an explicit split. The high-frequency branch uses depthwise separable convolutions with dual Swish-Tanh activation. The low-frequency branch applies window-based average pooling and lightweight self-attention, which reduces cost from $O((HW)^2)$ to $O(HW{\cdot}H'W')$. Cross-pathway fusion combines the two through reciprocal channel attention,
\begin{equation}
O_{\mathrm{fused}} = \mathrm{Sigmoid}\!\left(C_{\mathrm{high}}{\cdot}O_{\mathrm{low}} + C_{\mathrm{low}}{\cdot}O_{\mathrm{high}}\right)\odot X,
\label{eq:fuse}
\end{equation}
where $C_{\mathrm{high}}$ and $C_{\mathrm{low}}$ are softmax-normalized global pooling weights. DualFreqC3k2 replaces the C3k2 blocks at P4 and P5.

\subsection{Dilated Masked Attention for Scale-Calibrated Fusion}
\label{sec:c2dtab}
Because $\sigma$ ranges from 3 to 12 pixels, the $3\sigma$ envelope spans 9 to 36 pixels, so the fusion attention must match that range. C2DTAB combines a dilated grouped channel self-attention that models inter-feature dependencies with a masked window self-attention that samples sparsely through even-coordinate masking,
\begin{equation}
\mathrm{Mask}_{i,j} = \begin{cases}
0 & \text{if } (i \bmod 2 {=} 0)\wedge(j \bmod 2 {=} 0), \\
-\infty & \text{otherwise}.
\end{cases}
\label{eq:mask}
\end{equation}
This pattern aggregates peripheral context across the Gaussian envelope without the quadratic cost of dense attention. A dilated feed-forward network with rate $d{=}2$ expands the effective window from $3\times3$ to $7\times7$, which covers the $3\sigma$ range of 6 to 15\,mm nodules. C2DTAB replaces C2PSA in the neck.

\subsection{Scale-Adaptive Detection Heads}
\label{sec:ema}
\begin{figure}[t]
  \centering
  \includegraphics[width=0.94\linewidth]{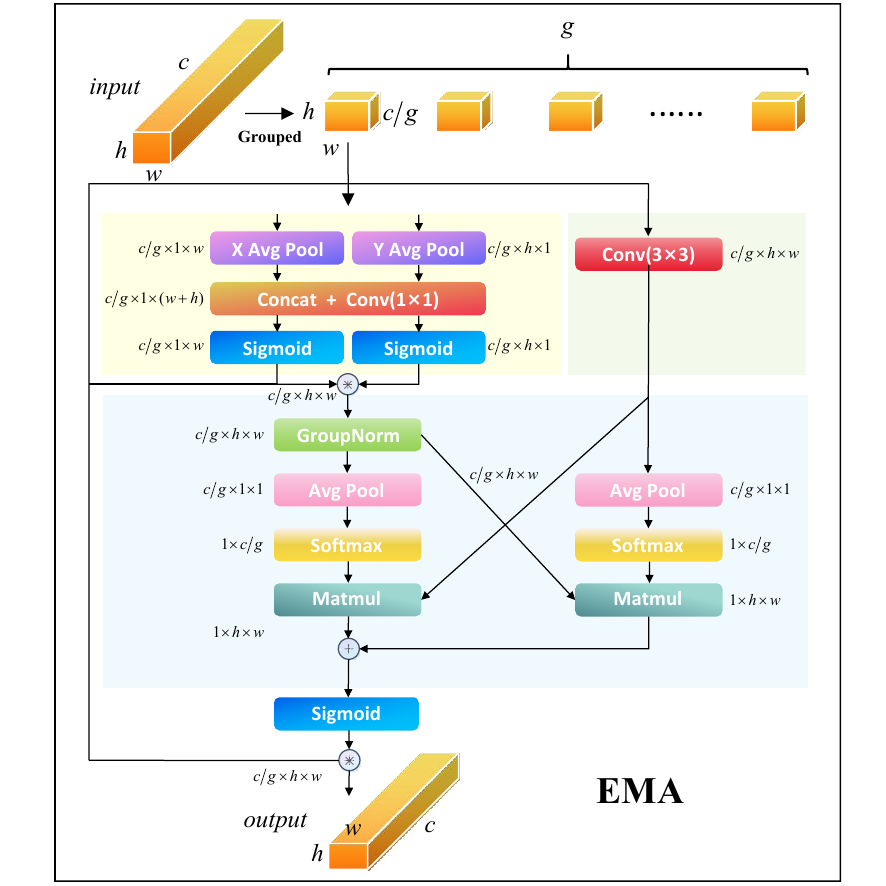}
  \caption{Efficient Multi-scale Attention head. Features are split into eight groups and reweighted by parallel spatial and convolutional pathways with softmax-normalized cross-pathway gating.}
  \label{fig:ema}
\end{figure}

Pulmonary lesions span from sub-6\,mm micro-nodules to tumors beyond 30\,mm, so the heads must allocate attention by scale. We place Efficient Multi-scale Attention~\cite{ouyang2023efficient} at P3 to P5. Each input is split into eight groups and processed by parallel spatial and convolutional pathways,
\begin{equation}
O_i = \sigma\!\left(C_s\,F_{\mathrm{conv}} + C_c\,W_{\mathrm{spatial}}\right)\odot X_i,
\label{eq:ema_head}
\end{equation}
where $C_s$ and $C_c$ are softmax-normalized weights and $\sigma$ is the sigmoid gate. At P3 the module emphasizes the local boundary detail of micro-nodules, and at P5 it prioritizes the global context that separates large masses from confluent vessels. Fig.~\ref{fig:ema} details the internal operations. The grouping preserves the orthogonal gradient components produced by PWConv, so directional sensitivity is not diluted during the attention step.

\subsection{Adaptive Loss with Dynamic Difficulty Tracking}
\label{sec:loss}
Sample difficulty evolves as the detector improves, and a fixed threshold saturates on easy samples while hard ones stay under-represented. The EWMASlide loss tracks the exponentially weighted moving average of prediction IoU with a threshold $\mu_t$,
\begin{equation}
\mu_t = d_t\,\mu_{t-1} + (1-d_t)\,\overline{\mathrm{IoU}}_t, \quad d_t = 0.999\left(1-e^{-t/2000}\right),
\label{eq:ewma}
\end{equation}
and assigns sample weights relative to that threshold,
\begin{equation}
w(\mathrm{IoU}) = \begin{cases}
1.0 & \mathrm{IoU}\le \mu_t-0.1, \\
e^{\,1-\mu_t} & \mu_t-0.1 < \mathrm{IoU} < \mu_t, \\
e^{-(\mathrm{IoU}-1)} & \mathrm{IoU}\ge \mu_t,
\end{cases}
\label{eq:slide}
\end{equation}
which induces a curriculum that shifts emphasis toward challenging samples. For regression we add the WiseIoU center-distance penalty~\cite{tong2023wise},
\begin{equation}
\mathcal{L}_{\mathrm{box}} = (1-\mathrm{IoU})\,e^{\,d^2/c^2},
\label{eq:wiou}
\end{equation}
where $d$ is the centroid distance and $c$ is the diagonal of the smallest enclosing box. This penalty suits nodules whose Gaussian peak defines a natural center. The unified objective sets $\lambda_{\mathrm{cls}}{=}1.0$ and $\lambda_{\mathrm{box}}{=}0.05$.

\section{Experiments}
\label{sec:exp}
\subsection{Datasets and Implementation}
We evaluate on three public benchmarks. KanserSet~\cite{model_api_dataset} contains 3{,}737 annotated images of solid, part-solid and ground-glass lesions from 3 to 28\,mm, split into 3{,}270, 311 and 156 for training, validation and testing. LUNA16~\cite{setio2017luna16,armato2011lung} comprises 888 scans with 1{,}186 nodules annotated by four radiologists, and preprocessing yields 3{,}558 samples split 70/15/15. Lung-PET-CT-Dx~\cite{li2020lungpetctdx} provides 13{,}295 images across four histological subtypes after balancing an original 102:1 ratio. Training runs on an NVIDIA RTX 4090 with PyTorch 2.1.0 using SGD at an initial rate of 0.01 with cosine annealing, momentum 0.937 and weight decay 0.0005, at $640\times640$ for KanserSet and $512\times512$ for the others, with early stopping at patience 100.

We constrain augmentation to operations that respect CT physics. Rotation and scaling preserve the radial symmetry of the prior and mimic changes in patient positioning and lesion size, and horizontal flipping produces anatomically valid mirror images. Brightness shifts within ten percent model the Hounsfield calibration differences across scanners, contrast scaling mimics different reconstruction kernels, and mild Gaussian noise approximates the quantum mottle of low-dose acquisition. We exclude mosaic augmentation, since it synthesizes anatomically impossible layouts and disrupts the contextual reasoning of the dilated attention.

\begin{table*}[t]
\centering
\caption{Comparison with state-of-the-art detectors on three public benchmarks. Blue, bold and underline together mark the best result per column within each dataset.}
\label{tab:comparison}
\setlength{\tabcolsep}{5pt}
\small
\begin{tabular}{@{}llcccccc@{}}
\toprule
\rowcolor{headrow}
\textbf{Dataset} & \textbf{Method} & \textbf{Precision} & \textbf{Recall} & \textbf{F1} & \textbf{mAP@0.5} & \textbf{mAP@0.75} & \textbf{mAP@0.5:0.95} \\
\midrule
\multirow{9}{*}{KanserSet}
& YOLOv8~\cite{yolov8_github}          & 0.945 & \best{0.925} & \best{0.935} & 0.928 & 0.762 & 0.651 \\
& YOLOv10~\cite{wang2024yolov10}       & 0.924 & 0.909 & 0.916 & 0.915 & 0.748 & 0.638 \\
& YOLOv11~\cite{yolov11_github}        & 0.933 & 0.804 & 0.864 & 0.879 & 0.721 & 0.612 \\
& Faster R-CNN~\cite{ren2016faster}    & 0.909 & 0.868 & 0.888 & 0.891 & 0.698 & 0.598 \\
& SSD~\cite{liu2016ssd}                & 0.892 & 0.818 & 0.853 & 0.862 & 0.672 & 0.571 \\
& MSDet~\cite{cai2024msdet}            & 0.938 & 0.912 & 0.925 & 0.931 & 0.790 & 0.660 \\
& Improved-YOLOv11-SSE~\cite{song2025improved} & 0.940 & 0.910 & 0.925 & 0.930 & 0.780 & 0.655 \\
& YOLOv5-CASP~\cite{zhang2023lung}     & 0.930 & 0.900 & 0.915 & 0.920 & 0.760 & 0.640 \\
\rowcolor{bestrow}
& \textbf{GRIPNet}                     & \best{0.965} & 0.899 & 0.931 & \best{0.953} & \best{0.892} & \best{0.726} \\
\midrule
\multirow{9}{*}{LUNA16}
& YOLOv8~\cite{yolov8_github}          & 0.855 & 0.846 & 0.850 & 0.875 & 0.712 & 0.601 \\
& YOLOv10~\cite{wang2024yolov10}       & 0.873 & 0.843 & 0.858 & 0.876 & 0.705 & 0.595 \\
& YOLOv11~\cite{yolov11_github}        & 0.882 & 0.842 & 0.862 & 0.897 & 0.721 & 0.608 \\
& Faster R-CNN~\cite{ren2016faster}    & 0.847 & 0.823 & 0.835 & 0.882 & 0.689 & 0.582 \\
& SSD~\cite{liu2016ssd}                & 0.836 & 0.792 & 0.813 & 0.854 & 0.665 & 0.561 \\
& MSDet~\cite{cai2024msdet}            & 0.895 & 0.860 & 0.877 & 0.905 & 0.745 & 0.630 \\
& Improved-YOLOv11-SSE~\cite{song2025improved} & 0.890 & 0.855 & 0.872 & 0.905 & 0.740 & 0.625 \\
& YOLOv5-CASP~\cite{zhang2023lung}     & 0.875 & 0.840 & 0.857 & 0.890 & 0.710 & 0.600 \\
\rowcolor{bestrow}
& \textbf{GRIPNet}                     & \best{0.903} & \best{0.878} & \best{0.890} & \best{0.916} & \best{0.775} & \best{0.655} \\
\midrule
\multirow{9}{*}{Lung-PET-CT-Dx}
& YOLOv8~\cite{yolov8_github}          & 0.943 & 0.897 & 0.919 & 0.935 & 0.678 & 0.571 \\
& YOLOv10~\cite{wang2024yolov10}       & 0.924 & 0.873 & 0.898 & 0.921 & 0.661 & 0.558 \\
& YOLOv11~\cite{yolov11_github}        & 0.931 & 0.907 & 0.919 & 0.951 & 0.689 & 0.581 \\
& Faster R-CNN~\cite{ren2016faster}    & 0.851 & 0.771 & 0.809 & 0.816 & 0.598 & 0.503 \\
& SSD~\cite{liu2016ssd}                & 0.872 & 0.781 & 0.824 & 0.820 & 0.612 & 0.515 \\
& MSDet~\cite{cai2024msdet}            & 0.955 & 0.940 & 0.947 & 0.962 & 0.690 & 0.595 \\
& Improved-YOLOv11-SSE~\cite{song2025improved} & 0.958 & 0.945 & 0.951 & 0.965 & 0.692 & 0.598 \\
& YOLOv5-CASP~\cite{zhang2023lung}     & 0.948 & 0.935 & 0.941 & 0.955 & 0.682 & 0.585 \\
\rowcolor{bestrow}
& \textbf{GRIPNet}                     & \best{0.968} & \best{0.963} & \best{0.965} & \best{0.979} & \best{0.702} & \best{0.613} \\
\bottomrule
\end{tabular}
\end{table*}

\subsection{Comparison with State-of-the-Art}
Table~\ref{tab:comparison} reports the comparison, which now includes three lung-nodule-specific detectors alongside general baselines. GRIPNet attains the highest mAP@0.5 on all three benchmarks, reaching 95.3\% on KanserSet against 93.1\% for the strongest competitor MSDet. The clearest separation appears at the stricter mAP@0.75 threshold, where GRIPNet reaches 89.2\% on KanserSet while every other method stays below 79\%. On LUNA16 GRIPNet leads all six metrics, and on Lung-PET-CT-Dx it reaches an mAP@0.5 of 97.9\% with an F1 of 96.5\%. The one metric where GRIPNet is not first is recall on KanserSet, where YOLOv8 is marginally higher at 92.5\% against 89.9\%; GRIPNet trades a little recall for a large gain in localization precision at high IoU, which matters more for reliable measurement of nodule extent.

The three lung-specific baselines behave as their designs predict. MSDet is the strongest competitor on KanserSet and Lung-PET-CT-Dx, which reflects its receptive-field enhancement for tiny nodules, yet it trails GRIPNet by 2.2 and 1.7 points of mAP@0.5. The gap widens at mAP@0.75, reaching 10.2 points over MSDet on KanserSet, because GRIPNet aligns box regression to the intensity peak rather than to a generic anchor. LUNA16 is the hardest setting, since multi-institutional acquisition lowers every method, and there GRIPNet still leads by 1.1 points of mAP@0.5 over both MSDet and the improved YOLOv11 while holding the best recall at 87.8\%. The improved YOLOv11 with spatial-SE attention is the closest rival on Lung-PET-CT-Dx, which is expected given its shared YOLOv11 backbone, but its fixed IoU penalty leaves a 1.4-point mAP@0.5 deficit that the adaptive curriculum of GRIPNet closes.

\subsection{Ablation Study}
Table~\ref{tab:ablation} adds the six components in sequence on KanserSet. The EMA-augmented baseline reaches an mAP@0.5 of 92.9\%. DualFreqC3k2 drops accuracy to 80.4\% while the two pathways adapt, and C2DTAB then recovers it to 88.3\% through a wider receptive field. PWConv marks the sharpest single gain and lifts mAP@0.5 to 92.0\%. The high-IoU behavior is instructive. The EWMASlide curriculum raises mAP@0.75 to 84.1\% and adds 4.8 points of recall, and combining it with the WiseIoU center penalty lifts the full model to 89.2\%. The center penalty alone is not enough, since configuration A+B+C+D+F without the curriculum falls to 67.8\% at mAP@0.75, so the localization gain comes from the curriculum and the center penalty acting together.

\begin{table}[t]
\centering
\caption{Ablation on KanserSet. Components: (A) EMA heads, (B) DualFreqC3k2, (C) C2DTAB, (D) PWConv, (E) EWMASlide loss, (F) WiseIoU. Blue, bold and underline together mark the best per column.}
\label{tab:ablation}
\setlength{\tabcolsep}{3pt}
\footnotesize
\begin{tabular}{@{}lcccccc@{}}
\toprule
\rowcolor{headrow}
\textbf{Config.} & \textbf{P} & \textbf{R} & \textbf{F1} & \textbf{AP$_{50}$} & \textbf{AP$_{75}$} & \textbf{AP$_{50:95}$} \\
\midrule
Baseline+A     & 0.955 & 0.820 & 0.882 & 0.929 & 0.748 & 0.624 \\
A+B            & 0.854 & 0.739 & 0.792 & 0.804 & 0.582 & 0.501 \\
A+B+C          & 0.900 & 0.809 & 0.852 & 0.883 & 0.799 & 0.620 \\
A+B+C+D        & 0.956 & 0.816 & 0.881 & 0.920 & 0.786 & 0.640 \\
A+B+C+D+E      & 0.943 & 0.864 & 0.902 & 0.932 & 0.841 & 0.654 \\
A+B+C+D+F      & 0.914 & 0.769 & 0.835 & 0.877 & 0.678 & 0.575 \\
\rowcolor{bestrow}
\textbf{Full GRIPNet} & \best{0.965} & \best{0.899} & \best{0.931} & \best{0.953} & \best{0.892} & \best{0.726} \\
\bottomrule
\end{tabular}
\end{table}

\subsection{Computational Efficiency}
Table~\ref{tab:eff} compares GRIPNet with the YOLOv11 baseline. The prior-guided modules raise parameters from 2.58\,M to 4.94\,M and GFLOPs from 6.3 to 8.5. This moderate overhead buys accuracy gains of 1.9 to 7.4 points in mAP@0.5, and throughput stays well above real time, from 230 to 309 FPS across the three benchmarks.

\begin{table}[t]
\centering
\caption{Efficiency versus the YOLOv11 baseline. FPS and mAP@0.5 are listed as KanserSet / LUNA16 / Lung-PET-CT-Dx.}
\label{tab:eff}
\setlength{\tabcolsep}{4pt}
\small
\begin{tabular}{@{}lcc@{}}
\toprule
\rowcolor{headrow}
\textbf{Metric} & \textbf{YOLOv11} & \textbf{GRIPNet} \\
\midrule
GFLOPs             & 6.3 & 8.5 \\
Parameters (M)     & 2.58 & 4.94 \\
Model size (MB)    & 5.3 & 9.8 \\
FPS                & 295 / 395 / 315 & 230 / 309 / 245 \\
\rowcolor{bestrow}
mAP@0.5            & 0.879 / 0.897 / 0.951 & \best{0.953 / 0.916 / 0.979} \\
\bottomrule
\end{tabular}
\end{table}

\subsection{Training Dynamics}
\begin{figure}[t]
  \centering
  \includegraphics[width=0.98\linewidth]{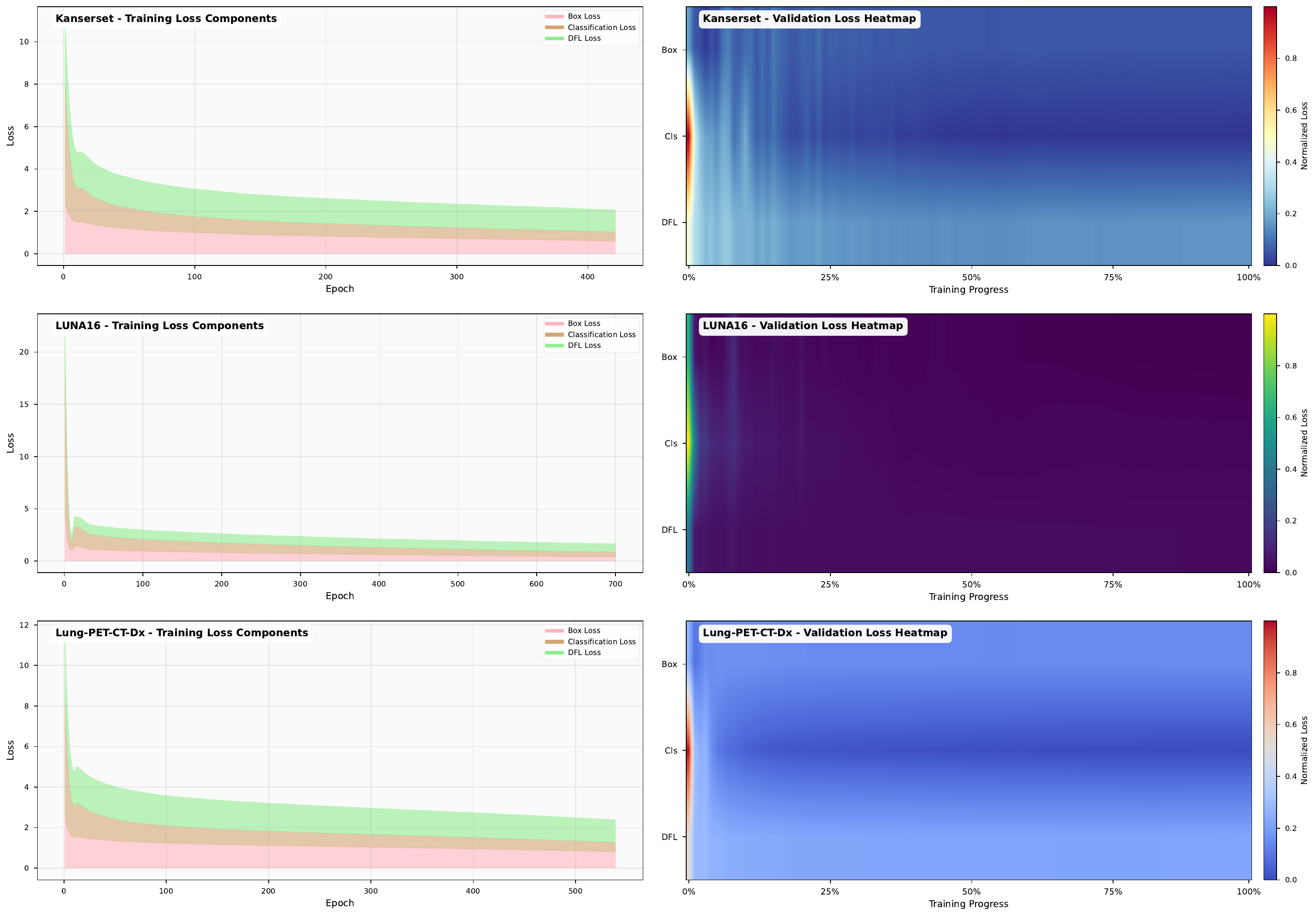}
  \caption{Training and validation loss on the three datasets. Each left panel plots the box, classification and DFL training losses against epoch, and each right panel is a normalized validation-loss heatmap of the same components over training progress.}
  \label{fig:loss}
\end{figure}

Fig.~\ref{fig:loss} tracks the training and validation loss on the three datasets. KanserSet stabilizes within about 100 epochs because its nodule presentations are relatively homogeneous. LUNA16 converges more gradually over roughly 266 epochs, which reflects the annotation spread of four-observer consensus. Lung-PET-CT-Dx reaches a stable plateau near epoch 200 after class balancing. The adaptive threshold of Eq.~\ref{eq:ewma} produces the mild oscillation visible early in training, since each rise of $\mu_t$ shifts optimization from easy samples toward harder ones before the representation settles. Convergence proceeds in two phases. Recall climbs first as the network locks onto the low-frequency Gaussian envelope, and precision then improves as the high-frequency pathway rejects vascular and pleural false positives.

\subsection{Qualitative Analysis}
\begin{figure*}[t]
  \centering
  \begin{subfigure}[b]{0.32\linewidth}
    \includegraphics[width=\linewidth]{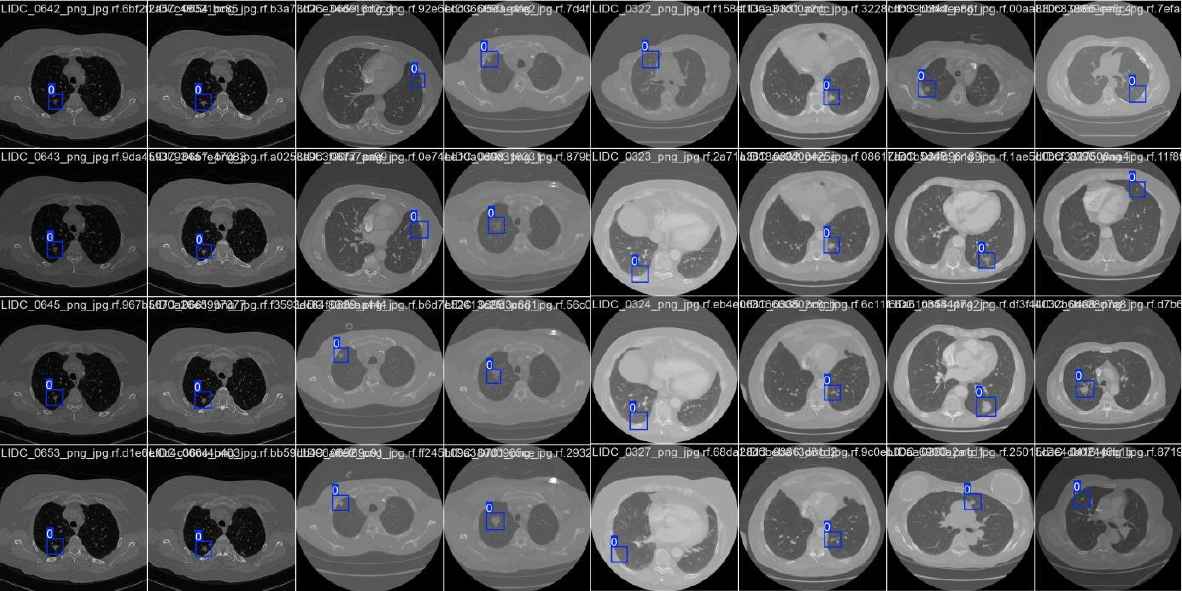}
    \caption{KanserSet}
  \end{subfigure}\hfill
  \begin{subfigure}[b]{0.32\linewidth}
    \includegraphics[width=\linewidth]{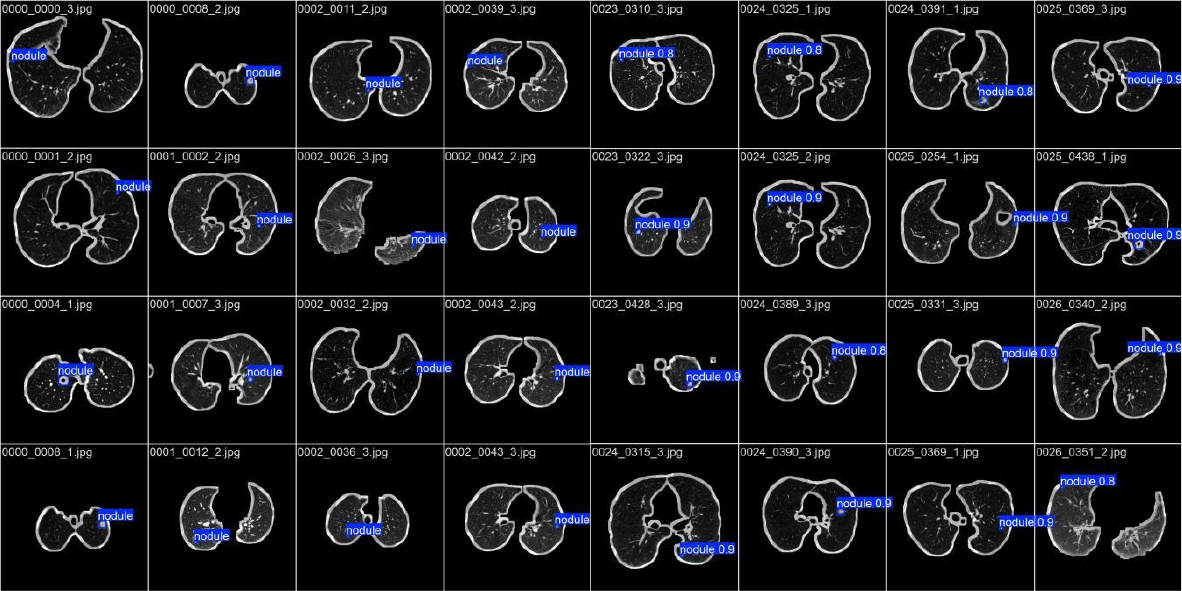}
    \caption{LUNA16}
  \end{subfigure}\hfill
  \begin{subfigure}[b]{0.32\linewidth}
    \includegraphics[width=\linewidth]{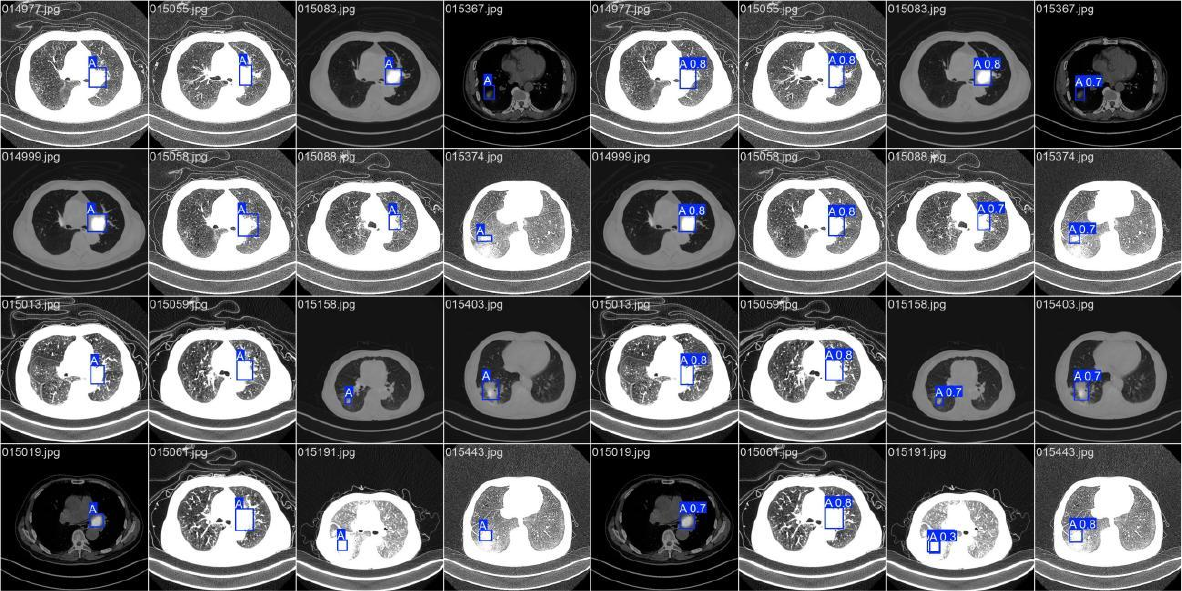}
    \caption{Lung-PET-CT-Dx}
  \end{subfigure}
  \caption{Representative detection results across the three benchmarks under diverse imaging conditions.}
  \label{fig:det}
\end{figure*}

Fig.~\ref{fig:det} shows representative detections. On KanserSet the predicted boxes conform tightly to sub-centimeter lesions where the gradient between periphery and parenchyma is subtle, which we attribute to the WiseIoU penalty anchoring regression to the Gaussian peak. On LUNA16 nodules near the pleura or vessels are delineated without spurious extension, a sign that the dilated masked attention constrains the receptive field to the decay envelope. On Lung-PET-CT-Dx the detector stays consistent across the four histological subtypes, consistent with the generality of a prior that captures the shared volumetric physics regardless of composition.

\subsection{Feature Map Analysis}
\begin{figure*}[t]
  \centering
  \begin{subfigure}[b]{0.88\linewidth}
    \includegraphics[width=\linewidth]{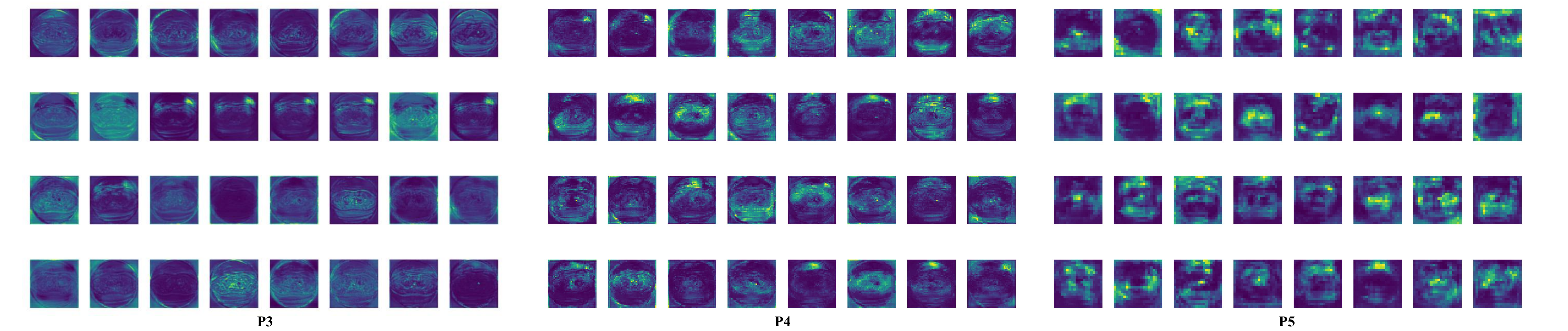}
    \caption{KanserSet}
    \label{fig:feat_k}
  \end{subfigure}
  \\[3pt]
  \begin{subfigure}[b]{0.88\linewidth}
    \includegraphics[width=\linewidth]{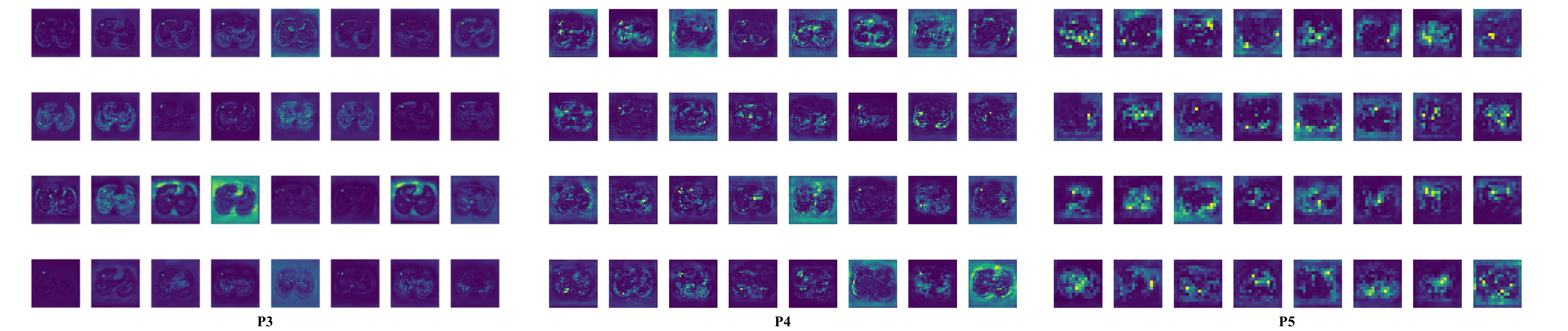}
    \caption{LUNA16}
    \label{fig:feat_luna}
  \end{subfigure}
  \\[3pt]
  \begin{subfigure}[b]{0.88\linewidth}
    \includegraphics[width=\linewidth]{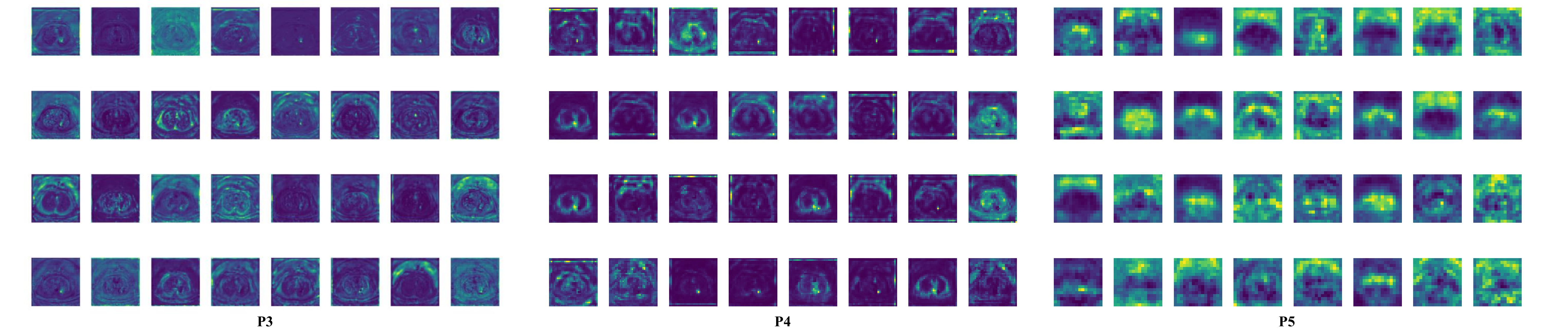}
    \caption{Lung-PET-CT-Dx}
    \label{fig:feat_dx}
  \end{subfigure}
  \caption{Feature maps at P3, P4 and P5 across the three benchmarks, each row running from P3 on the left to P5 on the right. P3 shows radial boundary activation, P4 adds peri-nodular context with visible frequency separation, and P5 responds to broader anatomical neighborhoods. The same progression holds on every dataset.}
  \label{fig:feat}
\end{figure*}

Fig.~\ref{fig:feat} visualizes multi-scale activations across the three benchmarks. At P3 the maps fire sharply along the nodule rim, often in radial directions, and the high resolution at this level preserves the detail that sub-6\,mm nodules with $\bar\sigma$ near 3 pixels demand. At P4 the response broadens to the peri-nodular ring predicted by the $3\sigma$ envelope, and the dual-frequency channels separate into a smooth interior group and a sharp group that isolates spiculation and lobulation. At P5 the maps track bronchi and vessels, the wider context that drives false-positive suppression. The same three-stage progression holds on all three datasets, and we read its meaning for module behavior and prior transfer in the discussion below.

\section{Discussion and Limitations}
The consistent gains across three datasets with different acquisition profiles indicate that the Gaussian radial prior transfers well, since it encodes the volumetric physics of X-ray attenuation rather than dataset-specific texture. The feature maps in Fig.~\ref{fig:feat} make this concrete. Activation moves with depth exactly as the prior predicts, from a radial rim response at P3, through the peri-nodular $3\sigma$ ring with clean dual-frequency separation at P4, to bronchial and vascular context at P5. The same three-stage progression appears on KanserSet, LUNA16 and Lung-PET-CT-Dx, so the network is not memorizing the texture of one cohort but tracking a physical pattern that all three share.

The detection examples in Fig.~\ref{fig:det} show the same logic at the output. On KanserSet the predicted boxes hug sub-centimeter lesions whose contrast against parenchyma is faint, which we read as the WiseIoU center penalty pulling regression toward the Gaussian peak rather than toward an arbitrary anchor. On LUNA16 the boxes stop cleanly at nodules that touch the pleura or a vessel, a sign that the dilated masked attention holds the receptive field to the decay envelope instead of bleeding into the adjacent wall. On Lung-PET-CT-Dx the detector behaves the same way across adenocarcinoma, small cell, large cell and squamous cell, which is what a prior grounded in shared volumetric physics rather than histology should do. The ablation in Table~\ref{tab:ablation} gives the mechanistic version of the story. The mAP@0.75 rise from 84.1 to 89.2 percent appears only when the EWMASlide curriculum and the center penalty act together, since neither alone clears 80 percent, so reliable high-IoU localization is a joint effect of learning which samples are hard and where their true center lies.

Two limits remain. GRIPNet operates on 2D slices, so a strongly anisotropic nodule can break the isotropic assumption behind Eq.~\ref{eq:gauss}, and a volumetric prior that fits an ellipsoidal Gaussian to the whole lesion is the natural next step. Recall on KanserSet also sits a little below YOLOv8, so the conspicuity-driven curriculum still trades a few borderline lesions for high-IoU precision, and a gentler difficulty schedule may recover them without loosening the boxes.

\section{Conclusion}
Progress in nodule detection has mostly come from larger backbones and wider search. We instead asked what a pulmonary nodule is, physically, and found a law rather than a texture. Across 18,218 lesions from three public benchmarks, intensity peaks at the geometric center and decays radially in a Gaussian pattern with a mean $R^2$ above 0.86, stable across every size stratum and histological subtype. GRIPNet turns that single measurement into an architecture, so pinwheel convolutions read the radial gradient, a dual-frequency split separates the smooth core from the spiculated rim, dilated masked attention matches the decay extent, and an adaptive loss tracks each lesion by its conspicuity. The detector reaches state-of-the-art accuracy at real-time speed, with an mAP@0.5 of 95.3\%, 91.6\% and 97.9\% and its widest margins in the strict high-IoU regime that governs reliable measurement. Because the prior encodes the physics of X-ray attenuation rather than the style of one cohort, it stays stable across scanners and folds and should extend to any roughly spherical lesion with a dense core, from hepatic to brain metastases.

\bibliographystyle{IEEEtran}
\bibliography{main}

@article{sung2021global,
  title={Global cancer statistics 2020: {GLOBOCAN} estimates of incidence and mortality worldwide for 36 cancers in 185 countries},
  author={Sung, Hyuna and Ferlay, Jacques and Siegel, Rebecca L and Laversanne, Mathieu and Soerjomataram, Isabelle and Jemal, Ahmedin and Bray, Freddie},
  journal={CA: A Cancer Journal for Clinicians},
  volume={71}, number={3}, pages={209--249}, year={2021}, publisher={Wiley}
}

@article{afridi2025minimally,
  title={Reduced lung-cancer mortality with volume {CT} screening in a randomized trial},
  author={de Koning, Harry J and van der Aalst, Carlijn M and de Jong, Pim A and others},
  journal={New England Journal of Medicine},
  volume={382}, number={6}, pages={503--513}, year={2020}
}

@misc{model_api_dataset,
  title={{KanserSet}: Pulmonary Nodule Detection Dataset},
  author={{Roboflow Community}},
  year={2023},
  howpublished={\url{https://universe.roboflow.com/}}
}

@article{setio2017luna16,
  title={Validation, comparison, and combination of algorithms for automatic detection of pulmonary nodules in computed tomography images: the {LUNA16} challenge},
  author={Setio, Arnaud Arindra Adiyoso and Traverso, Alberto and de Bel, Thomas and others},
  journal={Medical Image Analysis},
  volume={42}, pages={1--13}, year={2017}, publisher={Elsevier}
}

@article{armato2011lung,
  title={The lung image database consortium ({LIDC}) and image database resource initiative ({IDRI}): a completed reference database of lung nodules on {CT} scans},
  author={Armato III, Samuel G and McLennan, Geoffrey and Bidaut, Luc and others},
  journal={Medical Physics},
  volume={38}, number={2}, pages={915--931}, year={2011}
}

@misc{li2020lungpetctdx,
  title={A large-scale {CT} and {PET/CT} dataset for lung cancer diagnosis ({Lung-PET-CT-Dx})},
  author={Li, Pengyi and Wang, Shuo and Li, Tao and Lu, Jianfei and HuangFu, Yandong and Wang, Dongming},
  year={2020},
  howpublished={The Cancer Imaging Archive},
  doi={10.7937/TCIA.2020.NNC2-0461}
}

@inproceedings{ren2016faster,
  title={Faster {R-CNN}: towards real-time object detection with region proposal networks},
  author={Ren, Shaoqing and He, Kaiming and Girshick, Ross and Sun, Jian},
  booktitle={Advances in Neural Information Processing Systems},
  pages={91--99}, year={2015}
}

@inproceedings{liu2016ssd,
  title={{SSD}: single shot multibox detector},
  author={Liu, Wei and Anguelov, Dragomir and Erhan, Dumitru and Szegedy, Christian and Reed, Scott and Fu, Cheng-Yang and Berg, Alexander C},
  booktitle={European Conference on Computer Vision},
  pages={21--37}, year={2016}, publisher={Springer}
}

@article{terven2023comprehensive,
  title={A comprehensive review of {YOLO} architectures in computer vision: from {YOLOv1} to {YOLOv8} and {YOLO-NAS}},
  author={Terven, Juan and Cordova-Esparza, Diana-Margarita and Romero-Gonzalez, Julio-Alejandro},
  journal={Machine Learning and Knowledge Extraction},
  volume={5}, number={4}, pages={1680--1716}, year={2023}, publisher={MDPI}
}

@misc{yolov8_github,
  author={Jocher, Glenn and Chaurasia, Ayush and Qiu, Jing},
  title={{Ultralytics YOLOv8}},
  year={2023}, publisher={GitHub},
  howpublished={\url{https://github.com/ultralytics/ultralytics}}
}

@article{wang2024yolov10,
  title={{YOLOv10}: real-time end-to-end object detection},
  author={Wang, Ao and Chen, Hui and Liu, Lihao and Chen, Kai and Lin, Zijia and Han, Jungong and Ding, Guiguang},
  journal={arXiv preprint arXiv:2405.14458}, year={2024}
}

@misc{yolov11_github,
  author={Jocher, Glenn and Qiu, Jing},
  title={{Ultralytics YOLO11}},
  year={2024}, publisher={GitHub},
  howpublished={\url{https://github.com/ultralytics/ultralytics}}
}

@inproceedings{cai2024msdet,
  title={{MSDet}: receptive field enhanced multiscale detection for tiny pulmonary nodule},
  author={Cai, Guohui and Zhang, Ruicheng and He, Hongyang and Zhang, Zeyu and Ergu, Daji and Cao, Yuanzhouhan and Zhao, Jinman and Hu, Binbin and Liao, Zhibin and Zhao, Yang and Cai, Ying},
  booktitle={IEEE International Conference on Multimedia and Expo (ICME)},
  year={2025},
  note={arXiv:2409.14028}
}

@article{song2025improved,
  title={Improved {YOLO}-based pulmonary nodule detection with spatial-{SE} attention and an aspect ratio penalty},
  author={Song, Xinhang and Xie, Haoran and Gao, Tianding and Cheng, Nuo and Gou, Jianping},
  journal={Sensors},
  volume={25}, number={14}, pages={4245}, year={2025}, publisher={MDPI},
  doi={10.3390/s25144245}
}

@article{zhang2023lung,
  title={Lung nodule detection in medical images based on improved {YOLOv5s}},
  author={Zhang, Yi and others},
  journal={IEEE Access},
  volume={11}, pages={77254--77264}, year={2023}, publisher={IEEE},
  doi={10.1109/ACCESS.2023.3296530}
}

@article{halder2020lung,
  title={Lung nodule detection from feature engineering to deep learning in thoracic {CT} images: a comprehensive review},
  author={Halder, Amitava and Dey, Debangshu and Sadhu, Anup Kumar},
  journal={Journal of Digital Imaging},
  volume={33}, number={3}, pages={655--677}, year={2020}
}

@article{litjens2017survey,
  title={A survey on deep learning in medical image analysis},
  author={Litjens, Geert and Kooi, Thijs and Bejnordi, Babak Ehteshami and others},
  journal={Medical Image Analysis},
  volume={42}, pages={60--88}, year={2017}
}

@inproceedings{lin2017feature,
  title={Feature pyramid networks for object detection},
  author={Lin, Tsung-Yi and Doll{\'a}r, Piotr and Girshick, Ross and He, Kaiming and Hariharan, Bharath and Belongie, Serge},
  booktitle={CVPR}, pages={2117--2125}, year={2017}
}

@inproceedings{woo2018cbam,
  title={{CBAM}: convolutional block attention module},
  author={Woo, Sanghyun and Park, Jongchan and Lee, Joon-Young and Kweon, In So},
  booktitle={ECCV}, pages={3--19}, year={2018}
}

@inproceedings{ouyang2023efficient,
  title={Efficient multi-scale attention module with cross-spatial learning},
  author={Ouyang, Daliang and He, Su and Zhang, Guozhong and Luo, Mingzhu and Guo, Huaiyong and Zhan, Jian and Huang, Zhijie},
  booktitle={ICASSP}, pages={1--5}, year={2023}
}

@inproceedings{liu2021swin,
  title={Swin transformer: hierarchical vision transformer using shifted windows},
  author={Liu, Ze and Lin, Yutong and Cao, Yue and Hu, Han and Wei, Yixuan and Zhang, Zheng and Lin, Stephen and Guo, Baining},
  booktitle={ICCV}, pages={10012--10022}, year={2021}
}

@inproceedings{dosovitskiy2020image,
  title={An image is worth 16x16 words: transformers for image recognition at scale},
  author={Dosovitskiy, Alexey and Beyer, Lucas and Kolesnikov, Alexander and others},
  booktitle={ICLR}, year={2021}
}

@inproceedings{li2025rethinking,
  title={Rethinking transformer-based blind-spot network for self-supervised image denoising},
  author={Li, Junyi and Zhang, Zhilu and Zuo, Wangmeng},
  booktitle={AAAI}, year={2025}
}

@inproceedings{lin2017focal,
  title={Focal loss for dense object detection},
  author={Lin, Tsung-Yi and Goyal, Priya and Girshick, Ross and He, Kaiming and Doll{\'a}r, Piotr},
  booktitle={ICCV}, pages={2980--2988}, year={2017}
}

@article{tong2023wise,
  title={{Wise-IoU}: bounding box regression loss with dynamic focusing mechanism},
  author={Tong, Zanjia and Chen, Yuhang and Xu, Zewei and Yu, Rong},
  journal={arXiv preprint arXiv:2301.10051}, year={2023}
}

@article{raissi2019physics,
  title={Physics-informed neural networks: a deep learning framework for solving forward and inverse problems involving nonlinear partial differential equations},
  author={Raissi, Maziar and Perdikaris, Paris and Karniadakis, George E},
  journal={Journal of Computational Physics},
  volume={378}, pages={686--707}, year={2019}
}

@article{yang2025pinwheel,
  title={Pinwheel-shaped convolution and scale-based dynamic loss for infrared small target detection},
  author={Yang, Jiangnan and Liu, Shuangli and Wu, Jingjun and Su, Xinyu and Hai, Nan and Huang, Xueli},
  journal={arXiv preprint arXiv:2412.16986}, year={2024}
}

@book{motulsky2004fitting,
  title={Fitting Models to Biological Data Using Linear and Nonlinear Regression},
  author={Motulsky, Harvey J and Christopoulos, Arthur},
  year={2004}, publisher={Oxford University Press}
}

@article{yankelevitz1999small,
  title={Small pulmonary nodules: volumetrically determined growth rates based on {CT} evaluation},
  author={Yankelevitz, David F and Reeves, Anthony P and Kostis, William J and Zhao, Binsheng and Henschke, Claudia I},
  journal={Radiology},
  volume={217}, number={1}, pages={251--256}, year={2000}
}

\end{document}